\documentclass[11pt,a4paper]{article}
\usepackage[a4paper,margin=3.5cm]{geometry}
\usepackage{amscd,amssymb,amsthm,amsmath,comment}
\usepackage{apacite,lscape}

\begin{document}
\noindent{\Large \textbf{Understanding partition comparison indices\\ based on counting object pairs}}

\vspace{1cm}
\noindent\textbf{Matthijs J. Warrens}
\footnote{Matthijs J. Warrens: Groningen Institute for Educational Research, University of Groningen, the Netherlands (e-mail: m.j.warrens@rug.nl); Hanneke van der Hoef: Groningen Institute for Educational Research, University of Groningen, the Netherlands (e-mail: h.van.der.hoef@rug.nl).}
\\
\noindent\textbf{Hanneke van der Hoef}

\vspace{1cm}
\noindent\textbf{Abstract}\\
In unsupervised machine learning, agreement between partitions is commonly assessed with so-called external validity indices. Researchers tend to use and report indices that quantify agreement between two partitions for all clusters simultaneously. Commonly used examples are the Rand index and the adjusted Rand index. Since these overall measures give a general notion of what is going on, their values are usually hard to interpret.

Three families of indices based on counting object pairs are analyzed. It is shown that the overall indices can be decomposed into indices that reflect the degree of agreement on the level of individual clusters. The overall indices based on the pair-counting approach are sensitive to cluster size imbalance: they tend to reflect the degree of agreement on the large clusters and provide little to no information on smaller clusters. Furthermore, the value of Rand-like indices is determined to a large extent by the number of pairs of objects that are not joined in either of the partitions. 

\vspace{0.5cm}
\noindent\textbf{Key words}\\
Clustering comparison, external validity indices, reference standard partition, trial partition, Wallace indices, cluster size imbalance.

\vspace{0.5cm}
\noindent\textbf{MSC 2010 Subject classification}\\
62H30 Classification and discrimination; cluster analysis\\
62H20 Measures of association

\vspace{\fill}
\newpage
\section{Introduction}
The problem of measuring agreement between two different partitions of the same finite set of objects reappears continually in many scientific disciplines (Hubert 1977; Pfitzner, Leibbrandt and Powers 2009; Hennig, Meil\u{a}, Murtagh and Rocci 2015; Rezaei and Fr\"anti 2016). For example, in unsupervised machine learning, to evaluate the performance of a clustering method, researchers typically assess agreement between a reference standard partition that purports to represent the true cluster structure of the objects (golden standard), and a trial partition produced by the method that is being evaluated (Wallace 1983; Halkidi, Batiskis and Vazirgiannis 2002; Jain 2010). High agreement between the two partitions may indicate good recovery of the true cluster structure. 

Agreement between partitions can be assessed with so-called external validity indices (Albatineh, Niewiadomska-Bugaj and Mihalko 2006; Brun et al. 2007; Warrens 2008a, 2008b; Pfitzner et al. 2009). External validity indices can be roughly categorized into three approaches, namely 1) counting object pairs, 2) information theory (Vinh, Epps and Bailey 2010; Lei et al. 2016), and 3) matching sets (Rezaei and Fr\"anti 2016). Most external validity indices are of the pair-counting approach, which is based on counting pairs of objects placed in identical and different clusters. Information theoretic indices are based on concepts like the mutual information, Shannon entropy (Shannon 1948) and joint entropy (Kvalseth 1987; Pfitzner et al. 2009). These indices assess the difference in information between two partitions. Finally, set-matching indices are based on matching entire clusters, usually using the matched parts of each cluster, while ignoring the unmatched parts (Meil\u{a} 2007).

Commonly used external validity indices are the Rand index (Rand 1971) and the Hubert-Arabie adjusted Rand index (Hubert and Arabie 1985; Steinley 2004; Steinley, Brusco and Hubert 2016). Both these indices are based on counting pairs of objects. The adjusted Rand index corrects the Rand index for agreement due to chance (Albatineh et al. 2006; Warrens 2008a). Milligan and Cooper (1986), Milligan (1996), and Steinley (2004) proposed to use the adjusted Rand index as a standard tool in cluster validation research. However, the Rand index continues to be a popular validity index, probably because it has a simple, natural interpretation (Anderson, Bezdek, Popescu and Keller 2010).

Researchers tend to use and report validity indices that quantify agreement between two partitions for all clusters simultaneously (Milligan and Cooper 1986; Kim, Kim, Ashlock and Nam 2009; Albatineh and Niewiadomska-Bugaj 2011a; Yu, You, Wong and Han 2012; Alok, Saha and Ekbal 2014). Since these overall measures give a general notion of what is going on, it is usually difficult to pinpoint what their values, usually between 0 and 1, actually reflect. Values of overall indices are generally hard to interpret, except perhaps for values close to 0 or 1. 

In this paper we analyze three families of indices that are based on counting pairs of objects. We focus on indices based on pair-counting because these are most commonly used (Rezaei and Fr\"anti 2016). To enhance our understanding of overall indices, we show that various overall indices can be decomposed into indices that reflect the degree of agreement on the level of individual clusters. More precisely, we show that the overall indices are weighted means (variously defined) of indices that can be used to assess agreement for individual clusters of the partitions. In many cases the weights of these means are quadratic functions of the cluster sizes. 

The decompositions show that measures like the Jaccard index (Jaccard 1912) and the Hubert-Arabie adjusted Rand index (Hubert and Arabie 1985) tend to mainly reflect the degree of agreement between the partitions on the large clusters. The indices provide little to no information on the smaller clusters. This sensitivity to cluster size imbalance has been observed previously in the literature for some indices (Pfitzner et al. 2009; Vinh, Epps and Bailey 2009, 2010; Fr\"anti, Rezaei and Zhao 2014; Rezaei and Fr\"anti 2016). The analyses presented in this paper amplify these previous studies by providing insight into how this phenomenon actually works and to which indices it applies. Furthermore, the value of Rand-like indices is determined to a large extent by the number of pairs of objects that are not joined in either of the partitions.

The paper is organized as follows. The notation is introduced in Section 2. In Sections 3, 4 and 5 we present decompositions of three families of indices. Section 3 focuses on indices that are functions of the two asymmetric Wallace indices (Wallace 1983). Prototypical examples of this family are the Jaccard index and an index by Fowlkes and Mallows (1983). In Section 4 we analyze indices that are functions of both the Wallace indices and two indices that focus on pairs of objects that are not joined together in the partitions. A prototypical example of this family is the Rand index (Rand 1971). Decompositions of indices that are adjusted for agreement for chance (Albatineh et al. 2006; Warrens 2008a; Vinh et al. 2009) are presented in Section 5. A prototypical example of this family is the Hubert-Arabie adjusted Rand index. In Section 6 we present artificial and real-world examples to illustrate how the indices associated with the families in Sections 3 and 5 are related. In Section 7 we consider particular properties of the Rand-like family from Section 4. Finally, Section 8 contains a discussion.

\section{Notation}
In this section we introduce the notation. Suppose the data are scores of $n$ objects on $k$ variables. Let $U=\left\{U_1,U_2,\hdots,U_I\right\}$ and $Z=\left\{Z_1,Z_2,\hdots,Z_J\right\}$ denote two partitions of the objects, for example, a reference standard partition and a trial partition that was obtained with a clustering method that is being evaluated. Let $\textbf{N}=\left\{n_{ij}\right\}$ be a matching table of size $I\times J$ where $n_{ij}$ indicates the number of objects placed in cluster $U_i$ of the first partition and in cluster $Z_j$ of the second partition. The cluster sizes in respective partitions are the row and column totals of $\textbf{N}$, that is,
\begin{equation}
|U_i|=n_{i+}=\sum^{J}_{j=1}n_{ij}\qquad\mbox{and}\qquad|Z_j|=n_{+j}=\sum^{I}_{i=1}n_{ij}.
\end{equation} 
Following Fowlkes and Mallows (1983), the information in the matching table $\textbf{N}$ can be summarized in a fourfold contingency table (like Table 1) by counting several different types of pairs of objects: $N:=n(n-1)/2$ is the total number of pairs of objects,
\begin{equation}
T:=\sum^{I}_{i=1}\sum^{J}_{j=1}\binom{n_{ij}}{2}
\end{equation}
is the number of object pairs that were placed in the same cluster in both partitions, 
\begin{equation}
P:=\sum^{I}_{i=1}\binom{n_{i+}}{2}
\end{equation}
is the number of object pairs that were placed in the same cluster in partition $U$, and
\begin{equation}
Q:=\sum^{J}_{j=1}\binom{n_{+j}}{2}
\end{equation}
is the number of object pairs that were placed in the same cluster in partition $Z$. The bottom panel of Table 1 gives a representation of the matching table in terms of the counts $N$, $T$, $P$ and $Q$. 

Furthermore, define $a:=T$, $b:=P-T$, $c:=Q-T$ and $d:=N+T-P-Q$. Quantity $b$ ($c$) is the number of object pairs that were placed in the same cluster in partition $U$ ($Z$) but in different clusters in partition $Z$ ($U$). The quantity $d$ is the number of object pairs that are not joined in either of the partitions. The top panel of Table 1 gives a representation of the matching table using the counts $a$, $b$, $c$ and $d$. The latter notational system is commonly used for expressing similarity measures for $2\times 2$ tables (Warrens 2008b, 2008c, 2019; Heiser and Warrens 2010).

\begin{table}[t]
\caption{Two $2\times 2$ contingency table representations of matching table $\textbf{N}$.}
\begin{center}
\begin{tabular}{lccc}\hline
First partition &\multicolumn{3}{c}{Second partition}\\\cline{2-4}
 & Pair in the  & Pair in different & Totals\\
Representation 1 & same cluster & clusters & \\\hline
Pair in the same cluster   & $a$ & $b$ & $a+b$\\
Pair in different clusters & $c$ & $d$ & $c+d$\\
Totals                   & $a+c$ & $b+d$ & $N$\\\hline\\
Representation 2&\\\hline
Pair in the same cluster   & $T$   & $P-T$     & $P$\\
Pair in different clusters & $Q-T$ & $N+T-P-Q$ & $N-P$\\
Totals                     & $Q$   & $N-Q$     & $N$\\\hline
\end{tabular}\end{center}
\end{table}

Several authors have proposed indices specifically for assessing agreement between partitions (Rand 1971; Fowlkes and Mallows 1983; Wallace 1983; Hubert and Arabie 1985). However, if the agreement between the partitions is summarized as in the top panel of Table 1, one may use any similarity index from the vast literature on $2\times 2$ tables (Hub\'alek 1982; Baulieu 1989; Albatineh et al. 2006; Warrens 2008b, 2008c, 2019; Pfitzner et al. 2009). Moreover, each index that has been specifically proposed for assessing agreement between partitions, has a precursor in the literature on $2\times 2$ tables (see Tables 2, 3 and 4 for specific examples).

\section{Functions of the Wallace indices}
Wallace (1983) considers the following two asymmetric indices. The first index 
\begin{equation}
W=\frac{T}{P}=\frac{a}{a+b}
\end{equation} 
is the proportion of object pairs in the first partition that are also joined in the second partition (Severiano, Pinto, Ramirez and Carriço 2011). The second index 
\begin{equation}
V=\frac{T}{Q}=\frac{a}{a+c}
\end{equation} 
is the proportion of object pairs in the second partition that are also joined in the first partition. Table 2 presents twelve examples of indices from the literature that are increasing functions of conditional probabilities (5) and (6). 

Some of these functions, for example, the Dice index (Dice 1945)
\begin{equation}
D=\frac{2WV}{W+V}=\frac{2T}{P+Q},
\end{equation}
which is the harmonic mean of (5) and (6), are rather simple functions of the Wallace indices (e.g. sum, product, geometric mean, arithmetic mean, minimum, maximum), while other functions, for example the Jaccard coefficient, are more complicated functions of (5) and (6). Table 2 is a list of partition comparison indices that are functions of both $W$ and $V$. The middle column of Table 2 gives the formulas in terms of the regular $2\times 2$ tables. The last column of Table 2 gives the formula in terms of $W$ and $V$. All indices in Table 2 are increasing functions of $W$ and $V$. Hence, to understand all indices in Table 2, it is instrumental to first understand the values produced by indices (5) and (6).

\begin{landscape}
\begin{table}[ht]
\caption{Indices that are increasing functions of Wallace indices (5) and (6).}
\begin{center}
\begin{tabular}{lclcl}\hline
Source && Formula 1 && Formula 2\\\hline
Jaccard (1912)                                &&$a/(a+b+c)$          && $WV/(W+V-2WV)$\\
Gleason (1920), Dice (1945), S\o renson (1948)&&$2a/(2a+b+c)$        && $2WV/(W+V)$\\
Kulczy\'nski (1927), Driver and Kroeber (1932)&&$a/2(a+b)+a/2(a+c)$  && $(W+V)/2$\\  
Braun-Blanquet (1932)                         &&$a/(a+\max(b,c))$    && $\min(W,V)$\\
Simpson (1943)                                &&$a/(a+\min(b,c))$    && $\max(W,V)$\\
Ochiai (1957), Fowlkes and Mallows (1983)     &&$a/\sqrt{(a+b)(a+c)}$&& $\sqrt{WV}$\\
Sorgenfrei (1958), Cheetham and Hazel (1969)  &&$a^2/(a+b)(a+c)$     && $WV$\\
Sokal and Sneath (1963)                       &&$a/(a+2b+2c)$        && $WV/(2(W+V)-3WV)$\\
McConnaughey (1964)                           &&$(a^2-bc)/(a+b(a+c)$ && $W+V-1$\\
Johnson (1967)                                &&$a/(a+b)+a/(a+c)$    && $W+V$\\
Van der Maarel (1969)                         &&$(2a-b-c)/(2a+b+c)$  && $4WV/(W+V)-1$\\
Legendre and Legendre (1998)                  &&$3a/(3a+b+c)$        && $3WV/(W+V+WV)$\\\hline
\end{tabular}\end{center}
\end{table}
\end{landscape}

The Wallace indices can be decomposed into the following indices for the individual clusters of partitions $U$ and $Z$. Define for $U_i\in U$ the (relative) weights
\begin{equation}
P_i:=\binom{n_{i+}}{2}\qquad\mbox{and}\qquad p_i:=\frac{P_i}{P},
\end{equation}
which are, respectively, the number and proportion of object pairs in cluster $U_i$, and the index
\begin{equation}
w_i:=\sum\limits^J_{j=1}\binom{n_{ij}}{2}\Big/\binom{n_{i+}}{2},
\end{equation}
which is the proportion of object pairs in cluster $U_i$ that are joined in partition $Z$. Furthermore, define for $Z_j\in Z$ the (relative) weights
\begin{equation}
Q_j:=\binom{n_{+j}}{2}\quad\mbox{and}\quad q_j:=\frac{Q_j}{Q},
\end{equation}
which are, respectively, the number and proportion of object pairs in cluster $Z_j$, and the quantity
\begin{equation}
v_j:=\sum\limits^I_{i=1}\binom{n_{ij}}{2}\Big/\binom{n_{+j}}{2},
\end{equation}
which is the proportion of object pairs in cluster $Z_j$ that are joined in partition $U$. 

Indices (9) and (11) can be used to assess the agreement between partitions $U$ and $Z$ on the level of the individual clusters. Index (9) (or (11)) has value 1 if all objects in cluster $U_i$ ($Z_j$) are in precisely one cluster of partition $Z$ ($U$), and value 0 only if no two objects from cluster $U_i$ ($Z_j$) are paired together in partition $Z$ ($U$). Index (9) is a measure of sensitivity (recall, classification rate; Ting, 2011) that does not require any matching between clusters from partitions $U$ and $Z$.

We have the following decomposition for the first Wallace index. Index (5) is a weighted average of the indices in (9) using the $P_i$'s (or $p_i$'s) as weights:
\begin{equation}
W=\frac{\sum\limits^I_{i=1}w_iP_i}{\sum\limits^I_{i=1}P_i}=\sum\limits^I_{i=1}w_ip_i.
\end{equation}
Decomposition (12) shows that the overall $W$ value will, for a large part, be determined by the $w_i$ values of the clusters with high $P_i$ values, that is, the large clusters, since each $P_i$ is a quadratic function of the cluster size. The overall $W$ value will be high if, for each large cluster, its corresponding objects are assigned to the same cluster of partition $Z$, regardless of the $w_i$ values associated with smaller clusters. 

Furthermore, we have the following decomposition for the second Wallace index. Index (6) is a weighted average of the indices in (11) using the $Q_j$'s (or $q_j$'s) as weights:
\begin{equation}
V=\frac{\sum\limits^J_{j=1}w_jQ_j}{\sum\limits^J_{j=1}Q_j}=\sum\limits^J_{j=1}w_jq_j.
\end{equation}
Similarly, decomposition (13) shows that the overall $V$ value will, for a large part, be determined by the $v_j$ values of the clusters with high $Q_j$ values, that is, the large clusters. The overall $V$ value will be high if, for each large cluster, its corresponding objects are put in the same cluster of partition $U$. 

Decompositions (12) and (13) show that the indices in Table 2 are functions of the $w_i$'s and $v_j$'s of the individual clusters. Their values are largely determined by the $w_i$ values and $v_j$ values associated with the large clusters. For example, the Dice index is simply a weighted average of the $w_i$'s and $v_j$'s, using the $P_i$'s and $Q_j$'s as weights: 
\begin{equation}
D=\frac{\sum\limits^I_{i=1}w_iP_i+\sum\limits^J_{j=1}w_jQ_j}{\sum\limits^I_{i=1}P_i+\sum\limits^J_{j=1}Q_j}.
\end{equation}
The decompositions in (12), (13), and (14) are further explored with numerical examples in Section 6.

\section{Rand-like indices}
In addition to Wallace indices (5) and (6), we may consider the following two asymmetric indices. The first index 
\begin{equation}
W^*=\frac{N+T-P-Q}{N-P}=\frac{d}{c+d}.
\end{equation} 
is the proportion of object pairs not placed together in partition $Z$ that are also not joined in partition $U$. The second index 
\begin{equation}
V^*=\frac{N+T-P-Q}{N-Q}=\frac{d}{b+d}
\end{equation} 
is the proportion of object pairs not placed together in partition $U$ that are also not joined in partition $Z$. The quantity $N+T-P-Q$ in the numerator of (15) and (16) is the number of pairs that are not joined in either of the partitions. As an indication of agreement between the partitions, this quantity is rather neutral, counting pairs that are not clearly indicative of agreement (Wallace 1983). 

Table 3 presents eight examples of indices that are increasing functions of the four conditional probabilities (5), (6), (15), and (16). For example, the well-known Rand index (Rand 1971) is given by
\begin{equation}
R=\frac{N+2T-P-Q}{N}=\frac{a+d}{a+b+c+d}.
\end{equation}
The Rand index is a weighted average of indices (5), (6), (15) and (16), using the denominators of the indices as weights:
\begin{equation}
R=\frac{WP+VQ+W^*(N-P)+V^*(N-Q)}{P+Q+N-P+N-Q}.
\end{equation}
Furthermore, combining (18) with (12) and (13) we have the decomposition
\begin{equation}
R=\frac{\sum\limits^I_{i=1}w_iP_i+\sum\limits^J_{j=1}v_jQ_j+W^*(N-P)+V^*(N-Q)}{\sum\limits^I_{i=1}P_i+\sum\limits^J_{j=1}Q_j+2N-P-Q}.
\end{equation}
The decomposition in (19) shows that the Rand index can also been seen as a weighted average of the $w_i$'s, $v_j$'s and $W^*$ and $V^*$, using the $P_i$'s, $Q_j$'s and $(N-P)$ and $(N-Q)$ as weights.

\begin{landscape}
\begin{table}[ht]
\caption{Indices that are increasing functions of (5), (6), (15) and (16).}
\begin{center}
\begin{tabular}{lclcl}\hline
Source && Formula 1 && Formula 2\\\hline
Sokal and Michener (1958), Rand (1971) &&$(a+d)/(a+b+c+d)$                 && $R=$ formula (17)\\
Rogers and Tanimoto (1960)             &&$(a+d)/(a+2b+2c+d)$               && $R/(2-R)$\\
Hamann (1961)                          &&$(a-b-c+d)/(a+b+c+d)$             && $2R-1$\\
Sokal and Sneath (1963)                &&$2(a+d)/(2a+b+c+2d)$              && $2R/(R+1)$\\
Sokal and Sneath (1963)                &&$ad/\sqrt{(a+b)(a+c)(b+d)(c+d)}$  && $\sqrt{WVW^*V^*}$ \\
Sokal and Sneath (1963)                &&$a/4(a+b)+a/4(a+c)+$              && $(W+V+W^*+V^*)/4$\\
                                       &&$d/4(b+d)+d/4(c+d)$\\
Rogot and Goldberg (1966)              &&$a/(2a+b+c)+d/(b+c+2d)$           && $WV/(W+V)+W^*V^*/(W^*+V^*)$\\
Warrens (2008c)                        &&$4ad/(4ad+(a+d)(b+c))$            && $4/(W^{-1}+V^{-1}+W^{*-1}+V^{*-1})$ \\\hline
\end{tabular}\end{center}
\end{table}
\end{landscape}

\section{Chance-corrected functions}
Most indices from the literature have value 1 if there is perfect agreement between the two partitions. However, for many indices it is unclear under which conditions their theoretical lower bound, for example 0, is attained. Therefore, when partitions are compared, it is usually convenient that the index of choice has value 1 if the partitions are completely similar and value 0 if the partitions are statistically independent. For example, the Wallace indices in (5) and (6) have value 1 if the partitions are identical. However, their value is not necessarily 0 under statistical independence. 

If a similarity measure $S$ does not have value 0 under statistical independence, it can be corrected for agreement due to chance using the formula
\begin{equation}
AS=\frac{S-E(S)}{1-E(S)},
\end{equation}
where expectation $E(S)$ is conditional upon fixed row and column totals of matching table $\textbf{N}$, and 1 is the maximum value of $S$ regardless of the marginal numbers (Albatineh et al. 2006; Warrens 2008a; Vinh et al. 2009; Albatineh and Niewiadomska-Bugaj 2011b). 

Assuming a generalized hypergeometric model for matching table $\textbf{N}$, we have the expectation (Fowlkes and Mallows 1983; Hubert and Arabie 1985) 
\begin{equation}
E\binom{n_{ij}}{2}=\frac{1}{N}\binom{n_{i+}}{2}\binom{n_{+j}}{2}.
\end{equation}
Summing identity (21) over all cells of $\textbf{N}$ we obtain 
\begin{equation}
E\left(T\right)=\frac{PQ}{N}.
\end{equation}
Using Wallace index (5) in (20), together with identity (22), yields the adjusted index (Severiano et al. 2011)
\begin{equation}
AW=\frac{NT-PQ}{P(N-Q)}=\frac{ad-bc}{(a+b)(b+d)}. 
\end{equation}
Furthermore, inserting Wallace index (6) into (20) yields
\begin{equation}
AV=\frac{NT-PQ}{Q(N-P)}=\frac{ad-bc}{(a+c)(c+d)}. 
\end{equation}

\begin{landscape}
\begin{table}[ht]
\caption{Indices that are increasing functions of (23) and (24).}
\begin{center}
\begin{tabular}{lclcl}\\
Source && Formula 1 && Formula 2\\\hline
Doolittle (1885)                        &&$(ad-bc)^2/(a+b)(a+c)(b+d)(c+d)$          && $AW\cdot AV$\\
Yule (1912) (phi coefficient)           &&$(ad-bc)/\sqrt{(a+b)(a+c)(b+d)(c+d)}$     && $\sqrt{AW\cdot AV}$\\
Loevinger (1947)                        &&$(ad-bc)/\min[(a+b)(b+d),(a+c)(c+d)]$     && $\max(AW,AV)$\\  
Cohen (1960), Hubert and Arabie (1985)  &&$2(ad-bc)/[(a+b)(b+d)+(a+c)(c+d)]$        && $(2AW\cdot AV)/(AW+AV)$\\
Fleiss (1975)                           &&$(ad-bc)/2(a+b)(b+d)+(ad-bc)/2(a+c)(c+d)$ && $(AW+AV)/2$\\\hline
\end{tabular}\end{center}
\end{table}
\end{landscape}

Table 4 presents five examples of indices from the literature that are increasing functions of adjusted indices (23) and (24). A well-known example is the adjusted Rand index (Cohen 1960; Hubert and Arabie 1985; Steinley 2004; Warrens 2008d; Steinley, Brusco and Hubert 2016)
\begin{equation}
AR=\frac{2(NT-PQ)}{N(P+Q)-2PQ}=\frac{2(ad-bc)}{(a+b)(b+d)+(a+c)(c+d)}.
\end{equation}

The adjusted Rand index in (25) is the harmonic mean of (23) and (24). If $NT\neq PQ$, we have $AW>AR>AV$ if $P<Q$ and $AW<AR<AV$ if $P>Q$. The adjusted Rand index is what we get if we use the Rand index in (17) in correction for chance formula (20). Moreover, the adjusted Rand index is also obtained if we use the Dice index in (7) in (20), that is $AR=AD$ (Albatineh et al. 2006).

Indices (23) and (24) can be decomposed into the following indices for the individual clusters of partitions $U$ and $Z$. Using (9) in (20) we obtain 
\begin{equation}
Aw_i=\frac{N\sum\limits^J_{j=1}\displaystyle\binom{n_{ij}}{2}-\displaystyle\binom{n_{i+}}{2}Q}{\displaystyle\binom{n_{i+}}{2}(N-Q)}.
\end{equation}
Furthermore, inserting (11) into (20) yields
\begin{equation}
Av_j=\frac{N\sum\limits^I_{i=1}\displaystyle\binom{n_{ij}}{2}-\displaystyle\binom{n_{+j}}{2}P}{\displaystyle\binom{n_{+j}}{2}(N-P)}.
\end{equation}
Similar to indices (9) and (11), indices (26) and (27) can be used to assess the agreement between partitions $U$ and $Z$ on the level of the individual clusters. Index (26) (or (27)) has value 1 if all objects in cluster $U_i$ ($Z_j$) are in precisely one cluster of $Z$ ($U$), and value 0 under statistical independence. Index (26) is a measure of sensitivity (recall, classification rate) that does not require any matching between clusters from partitions $U$ and $Z$.

Index (23) is a weighted average of the indices in (26) using the $P_i$'s (or $p_i$'s) as weights:
\begin{equation}
AW=\frac{\sum\limits^I_{i=1}Aw_iP_i}{\sum\limits^I_{i=1}P_i}=\sum\limits^I_{i=1}Aw_ip_i.
\end{equation}
Decomposition (28) shows that the overall $AW$ value will for a large part be determined by the $Aw_i$ values of the clusters with high $P_i$ values, that is, the large clusters, since each $P_i$ is a quadratic function of the cluster size. The overall $AW$ value will be high if, for each large cluster, its corresponding objects are assigned to the same cluster of the second partition, regardless of the $Aw_i$ values associated with smaller clusters. 

Furthermore, index (24) is a weighted average of the indices in (27) using the $Q_j$'s (or $q_j$'s) as weights:

\begin{equation}
AV=\frac{\sum\limits^J_{j=1}Aw_jQ_j}{\sum\limits^J_{j=1}Q_j}=\sum\limits^J_{j=1}Aw_jq_j.
\end{equation}
Similarly, decomposition (29) shows that the overall $AV$ value will for a large part be determined by the $Av_j$ values of the clusters with high $Q_j$ values, that is, the large clusters. The overall $AV$ value will be high if, for each large cluster, its corresponding objects are put in the same cluster of the first partition. 

Decompositions (28) and (29) show that the adjusted Rand index is simply a weighted average of the $Aw_i$'s and $Av_j$'s, using the $P_i$'s and $Q_j$'s as weights: 
\begin{equation}
AR=\frac{\sum\limits^I_{i=1}Aw_iP_i+\sum\limits^J_{j=1}Aw_jQ_j}{\sum\limits^I_{i=1}P_i+\sum\limits^J_{j=1}Q_j}.
\end{equation}
The value of the adjusted Rand index will for a large part be determined by the $Aw_i$ values and $Av_j$ values corresponding to large clusters.

\section{Numerical examples}
In this section, we present examples to illustrate how the building blocks in (9) and (11) are related to the Wallace indices in (12) and (13), and how the building blocks in (26) and (27) are related to the adjusted Wallace indices in (23) and (24). We first consider four toy examples. In addition, we consider data on E. coli sequences (Horton \& Nakai, 1996; Lichman, 2013). The Rand-like indices from Section 4 are further considered in the next section.

For the first two toy examples, suppose we have two partitions of $n=56$ objects that both consist of 2 large clusters of size $n=20$, numbered $i,j\in\left\{1,2\right\}$, and 2 small clusters of size $n=8$, numbered $i,j\in\left\{3,4\right\}$. As a first toy example, suppose that there is perfect agreement between the partitions on the large clusters, whereas the small clusters are completely (uniformly) mixed up. In this case, we have $w_i=v_j=Aw_i=Av_j=1$ for $i,j\in\left\{1,2\right\}$ (large clusters), and $w_i=v_j=.43$ and $Aw_i=Av_j=.20$ for $i,j\in\left\{3,4\right\}$ (small clusters). Furthermore, the relative weights are $p_i=q_j=.44$ for $i,j\in\left\{1,2\right\}$ (large clusters), and $p_i=q_j=.06$ for $i,j\in\left\{3,4\right\}$ (small clusters). Moreover, the overall Wallace indices are $W=V=.93$, the Dice index is $D=.93$, the Rand index is $R=.96$, the adjusted Wallace indices are $AW=AV=.90$ and the adjusted Rand index is $AR=.90$. Thus, the overall values primarily reflect the perfect agreement on the two large clusters.

As a second toy example, suppose that the large clusters are completely (uniformly) mixed up, whereas there is perfect agreement on the small clusters. In this case we have $w_i=v_j=.47$ and $Aw_i=Av_j=.27$ for $i,j\in\left\{1,2\right\}$ (large clusters), and $w_i=v_j=Aw_i=Av_j=1$ for $i,j\in\left\{3,4\right\}$ (small clusters). Because the cluster sizes are identical to the cluster sizes in the first example, the relative weights are also identical to the ones from the first example. Moreover, the overall indices are $D=W=V=.54$, $R=.74$ and $AR=AW=AV=.36$. Thus, except for the Rand index ($R=.74$) the overall values primarily reflect the disagreement on the large clusters. 

The sensitivity to cluster size imbalance becomes more explicit when the clusters vary more in size. For the third and fourth example, suppose we have two partitions of $n=1056$ objects that both consist of 4 large clusters of size $n=200$, numbered $i,j\in\left\{1,2,3,4\right\}$, and 16 small clusters of size $n=16$, numbered $i,j\in\left\{5,6,\hdots,16\right\}$. As a third toy example, suppose that there is perfect agreement between the partitions on the large clusters, whereas the small clusters are completely (uniformly) mixed up. In this case we have $w_i=v_j=Aw_i=Av_j=1$ for $i,j\in\left\{1,2,3,4\right\}$ (large clusters), and $w_i=v_j=0$ and $Aw_i=Av_j=-.17$ for $i,j\in\left\{5,6,\hdots,16\right\}$ (small clusters). Furthermore, the relative weights are $p_i=q_j=.04$ for $i,j\in\left\{1,2,3,4\right\}$ (large clusters), and $p_i=q_j<.001$ for $i,j\in\left\{5,6,\hdots,16\right\}$ (small clusters). Moreover, the overall indices are $W=V=D=.98$, $R=.99$ and $AW=AV=AR=.97$. Thus, the overall values primarily reflect the perfect agreement on the large clusters.

As a fourth toy example, suppose that the large clusters are completely (uniformly) mixed up, whereas there is perfect agreement on the small clusters. In this case, we have $w_i=v_j=.25$ and $Aw_i=Av_j=.12$ for $i,j\in\left\{1,2,3,4\right\}$ (large clusters), and $w_i=v_j=Aw_i=Av_j=1.00$ for $i,j\in\left\{5,6,\hdots,16\right\}$ (small clusters). Because the cluster sizes are identical to the cluster sizes in the third example, the relative weights are also identical to the ones in the third example. Moreover, the overall indices are $D=W=V=.26$, $R=.78$, and $AR=AW=AV=.14$. Thus, except for the Rand index ($R=.78$) the overall values primarily reflect the disagreement on the large clusters. Rand-like indices are further studied in the next section.

In the remainder of this section we consider a data set that contains information on E. coli sequences (Horton and Nakai, 1996; Lichman, 2013). The data consist of 336 proteins belonging to 8 classes (reference partition), which are the localization sites: cytoplasmic (cp), inner membrane without signal sequence (im), inner membrane lipoprotein (imL), inner membrane, cleavable signal sequence (imS), inner membrane proteins with an uncleavable signal sequence (imU), outer membrane (om), outer membrane lipoprotein (omL), and periplasmic (pp). For all proteins, 7 features were calculated from amino acid sequences. Table 5 presents the matching table of the reference partition and a $K$-means clustering (Steinley 2006; Jain 2010; Huo, Ding, Liu and Tseng 2016) of the E. coli sequences. All 7 features were used in the analysis. The number of clusters was set to $K=4$. The reference partition consists of 8 clusters, whereas the trial partition consists of 4 clusters. The row totals of Table 5 are the class sizes.

Table 6 presents the values of various indices and weights corresponding to the data in Table 5.
We first consider the row indices. Most of the cp proteins are grouped together ($w_1=.92$ and $Aw_1=.88$). Many of the im proteins are grouped together ($w_2=.79$ and $Aw_2=.69$). None of the imL and imS proteins are grouped together ($w_3=w_4=0$ and $Aw_3=Aw_4=-.47$). Most of the imU proteins are grouped together ($w_5=.94$ and $Aw_5=.92$). Many of the om proteins are grouped together ($w_6=.81$ and $Aw_6=.72$). All of the omL proteins are grouped together ($w_7=Aw_7=1$). 

\begin{table}[p]
\caption{Matching table of a reference partition and a $K$-means clustering of E. coli sequences (Horton \& Nakai, 1996; Lichman, 2013).}
\begin{center}\begin{tabular}{llccccc}\hline
\multicolumn{2}{c}{Reference partition}\qquad\mbox{}&\multicolumn{4}{c}{Trial partition}&\\
Proteins & &$Z_1$&$Z_2$&$Z_3$&$Z_4$&Totals\\\hline
cp  & $=U_1$ &   5 &   0 & 137 &   1 & 143   \\
im  & $=U_2$ &   8 &   0 &   1 &  68 &  77   \\
imL & $=U_3$ &   0 &   1 &   0 &   1 &   2   \\
imS & $=U_4$ &   1 &   0 &   0 &   1 &   2   \\
imU & $=U_5$ &   0 &   0 &   1 &  34 &  35   \\
om  & $=U_6$ &   2 &  18 &   0 &   0 &  20   \\
omL & $=U_7$ &   0 &   5 &   0 &   0 &   5   \\
pp  & $=U_8$ &  46 &   1 &   4 &   1 &  52   \\\hline
Totals      &&  62 &  25 & 143 & 106 & 336\\\hline
\end{tabular}\end{center}
\end{table}

\begin{table}[p]
\caption{Row, column and overall statistics for the data in Table 5.}
\begin{center}\begin{tabular}{cccccccccccl}\hline
\multicolumn{4}{c}{Row statistics}&&\multicolumn{4}{c}{Column statistics}&&\multicolumn{2}{c}{Overall indices}\\
$i$ & $w_i$ & $Aw_i$ & $p_i$ && $j$ & $v_j$ & $Av_j$ & $q_j$ && \\\hline
1	&	.92	&	 .88 & .67    && 1 & .57	& .41 & .11 && $W$ & .88 \\ 
2	&	.79	&	 .69 & .19    && 2 & .54	& .37 & .02 && $V$ & .75 \\
3	&	 0  &	-.47 & $<.001$&& 3 & .92	& .89 & .57 && $D$ & .81 \\
4	&	 0	&	-.47 & $<.001$&& 4 & .51	& .33 & .31 && $AW$& .83 \\\cline{6-9}
5	&	.94	&	 .92 & .04    &&   &      &     &     && $AV$& .65 \\
6	&	.81	&	 .72 & .01    &&   &      &     &     && $AR$& .73 \\\cline{11-12}
7	&	 1  &	 1   & .001   &&   \\
8	&	 .79&	 .68 & .09    && \\\hline
\end{tabular}\end{center}
\end{table}

\vspace{\fill}
\newpage

\noindent Finally, many of the pp proteins are grouped together ($w_8=.79$ and $Aw_8=.68$).

The overall indices $W=.88$ and $AW=.83$ reflect that many of the proteins from the same class are grouped together in the $K$-means clustering. The overall values are weighted averages of the cluster indices associated with the rows of Table 5. The $W$ value and $AW$ value are almost completely determined by the values of the cluster indices associated with the two large classes, the cp and im proteins ($p_1=.67$ and $p_2=.19$). The values of the indices associated with the five smallest classes (imL, imS, imU, om, and omL) are basically immaterial for the calculation of the values of $W$ and $AW$.

Next, we consider the column indices. Since there are 8 classes of proteins and the $K$-means clustering consists of only 4 clusters, the recovery of the cluster structure as represented in the reference partition cannot be perfect. That is, some of the protein classes will be lumped together in the same cluster. The indices associated with clusters $Z_1$, $Z_2$ and $Z_4$ reflect that the clusters contain more than one type of protein ($v_1=.57$, $v_2=.54$, $v_4=.51$, and $Av_1=.41$, $Av_2=.37$ and $Av_4=.33$). Furthermore, the indices associated with cluster $Z_3$ tell us that at least one of the protein classes was recovered rather well by the $K$-means clustering ($v_3=.92$ and $Av_3=.89$). 

The overall indices $V=.75$ and $AV=.65$ reflect that some proteins from different classes have been grouped together in the $K$-means clustering. The overall values are weighted averages of the cluster indices associated with the columns of Table 5. The $V$ value and $AV$ value are completely determined by the values of the cluster indices associated with the three large clusters $Z_1$, $Z_3$ and $Z_4$ ($q_1=.11$, $q_3=.57$ and $q_4=.31$). The value of the index associated with the smallest cluster ($Z_2$) is not relevant for the calculation of the values of $V$ and $AV$. 

Finally, the Dice index $D=.81$ and the adjusted Rand index $AR=.73$ are harmonic means of, respectively, $W=.88$ and $V=.75$, and $AW=.83$ and $AV=.65$. Compared to the ordinary arithmetic mean of two numbers, the harmonic mean puts a bit more emphasis on the smallest of the two numbers. Therefore, the values of $D$ and $AR$ lie between, respectively, the values of $W$ and $V$, and $AW$ and $AV$, and just a little bit closer to the overall indices $V$ and $AV$.    

In summary, the five data examples show that indices that belong to the families of indices based on the Wallace indices in (5) and (6) and the adjusted Wallace indices in (23) and (24)  are quite sensitive to cluster size imbalance. The overall indices tend to mainly reflect the degree of agreement between the partitions on the large clusters. They provide little to no information on the recovery of the smaller clusters.

\section{More on Rand-like indices}
Indices that belong to the families of indices based on the Wallace indices in (5) and (6) and the adjusted Wallace indices in (23) and (24) can be understood in terms of indices for individual clusters (see Sections 3 and 5, respectively). However, this is quite different for the family of indices from Section 4. These Rand-like indices are increasing functions of the Wallace indices in (5) and (6) as well as the asymmetric indices in (15) and (16). The Rand index, which is the prototypical example of this family, may be interpreted as the ratio of the number of object pairs placed together in a cluster in each of the two partitions and the number of object pairs assigned to different clusters in both partitions, relative to the total number of object pairs. Rand-like indices combine two sources of information, object pairs put together in both partitions, which is reflected in Wallace indices (5) and (6), and object pairs assigned to different clusters in both partitions, which is reflected in indices (15) and (16).

\begin{table}[ht]
\caption{Values of indices and weights for the five data examples from Section 6.}
\begin{center}\begin{tabular}{lccccc}\hline
&\multicolumn{4}{c}{Toy examples}&E. coli\\\cline{2-5}
Statistic &   1 &   2 &  3 & 4 & \\\hline
$W$       & .93 & .54 & .98 & .26 & .88 \\
$V$       & .93 & .54 & .98 & .26 & .75 \\
$W^*$     & .97 & .82 &1.00 & .87 & .89 \\
$V^*$     & .97 & .82 &1.00 & .87 & .95 \\
$R$       & .96 & .74 & .99 & .78 & .89 \\\hline
$P/N$     & .28 & .28 & .15 & .15 & .27 \\
$Q/N$     & .28 & .28 & .15 & .15 & .32 \\
$(N-P)/N$ & .72 & .72 & .85 & .85 & .73 \\
$(N-Q)/N$ & .72 & .72 & .85 & .85 & .68 \\\hline
\end{tabular}\end{center}
\end{table}

To understand what the values of Rand-like indices may reflect requires knowledge of how the two sources of information on object pairs contribute to the overall values of the indices. The above interpretation suggests that both sources may contribute equally. Results presented in Warrens and Van der Hoef (2018) show that this is not the case. In this paper it is shown how the Rand index (Rand 1971) is related to the four asymmetric indices (5), (6), (15) and (16). Warrens and Van der Hoef (2018) systematically varied artificial data examples. The results of their study can be summarized as follows. In many situations, including cases of high, medium and low agreement between the partitions, and statistical independence of the partitions, the number of object pairs assigned to different clusters in both partitions is (much) higher than the number of object pairs that are combined in both partitions.

Decomposition (18) shows that the Rand index is a weighted average of the indices $W$, $V$, $W^*$ and $V^*$ using, respectively, the quantities $P$, $Q$, $(N-P)$ and $(N-Q)$ as weights. The results of Warrens and Van der Hoef (2018) have two consequences: 1) the values of $W$ and $V$ are usually (much) smaller than the values of $W^*$ and $V^*$; 2) the values of $P$ and $Q$ are usually (much) smaller than the values of $(N-P)$ and $(N-Q)$. The second consequence implies that the value of the Rand index will in many cases for a large part be determined by the values of $W^*$ and $V^*$. Furthermore, together the two consequences imply that the Rand index will usually produce a high values, say between .70 and 1.00, because $(N-P)$ and $(N-Q)$, the weights associated with $W^*$ and $V^*$, will in general also be high. Since all Rand-like indices presented in Table 3 are increasing functions of $W$, $V$, $W^*$ and $V^*$, these indices will generally produce high values as well.

The results in Warrens and Van der Hoef (2018) can be illustrated with the data examples from the previous section. Table 7 gives the values of indices $W$, $V$, $W^*$, $V^*$ and $R$ and relative weights $P/N$, $Q/N$, $(N-P)/N$ and $(N-Q)/N$ for the five data examples from Section 6. In all examples the relative weights $P/N$ and $Q/N$ are much smaller than the relative weights $(N-P)/N$ and $(N-Q)/N$. Thus, in each example the value of the Rand index will be influenced more by the values of $W^*$and $V^*$ than by the values of the Wallace indices $W$ and $V$. Furthermore, in each example the values of $W^*$and $V^*$ are quite high.

In summary, the Rand-like indices tend to reflect how much object pairs have been assigned to different clusters in both partitions. A first consequence is that they will generally produce high values (say between .70 and 1). A second consequence is that cluster size imbalance is less of an issue for these indices. 

\section{Discussion}
For assessing agreement between two partitions researchers usually use and report overall measures that quantify agreement for all clusters simultaneously. Since overall indices only give a general notion of what is going on their values are often hard to interpret. In this paper we analyzed three families of indices that are based on counting pairs of objects. We presented decompositions of the overall indices into indices that reflect the degree of agreement on the level of individual clusters. The decompositions make explicit what the building blocks of the overall indices are and how they are weighted, and thus provide insight into what information the values of overall indices may reflect.

Indices that are based on the Wallace indices, for example, the Jaccard index and an index by Fowlkes and Mallows, or the adjusted Wallace indices, for example, the adjusted Rand index, are quite sensitive to cluster size imbalance. They tend to reflect the degree of agreement between the partitions on the large clusters only. They provide little to no information on the agreement on smaller clusters. This property can be useful for overall indices because the large clusters contain the most object pairs. However, the property may not be desirable in all situations, for example, if one wants to assess the recovery of small clusters, which may be the more interesting clusters.

A third family of indices consists of Rand-like indices. These indices can be decomposed into a row and a column index that reflect how many object pairs are put together in both partitions, and into a row and a column index that reflect how many object pairs are put in different clusters in both partitions. They tend to reflect how much object pairs have been assigned to different clusters in both partitions. They will generally produce high values (say, between .70 and 1). Moreover, cluster size imbalance is less of an issue for these indices.

Sensitivity to cluster size imbalance of various indices has previously been observed in the classification literature (De Souto et al. 2012; Rezaei and Fr\"anti 2016). The analyses presented in this paper add some details to these studies by providing insight into how this phenomenon actually works, and to which indices it applies. The various indices are weighted means of cluster indices, and it is this weighting that introduces the sensitivity to cluster size imbalance. Several authors have proposed indices that are not sensitive to cluster size imbalance (see, for example, Pfitzner et al. 2009; Fr\"anti, Rezaei and Zhao 2014). 

A negative property of the Rand index that has been noted in the classification literature is that its value concentrates in a small interval near the value 1 (Fowlkes and Mallows 1983; Meil\u{a} 2007). The analyses presented in Warrens and Van der Hoef (2018) and in this paper provide insight into how this property works. Furthermore, the analyses show that the property also applies to other Rand-like indices, that is, indices that are increasing functions of the same quantities as the Rand index.  

In this paper we focused on indices that are based on counting pairs of objects. This type of index, especially the adjusted Rand, is most commonly used. Some of the ideas presented in this paper can be applied to other types of partition comparison indices. For example, decompositions of various normalizations of the mutual information (Pfitzner et al. 2009) are presented in Van der Hoef and Warrens (2019). It turns out that these information theoretic indices are also sensitive to cluster size imbalance, but in a more complicated way than the indices based on the pair-counting approach. 

\vspace{\fill}
\newpage


\vspace{\fill}
\end{document}